\documentclass[letterpaper]{article} % DO NOT CHANGE THIS
\usepackage[preprint]{aaai2027}  % Preprint: suppress the AAAI copyright block
\usepackage[hyphens]{url}  % DO NOT CHANGE THIS
\usepackage{graphicx} % DO NOT CHANGE THIS
\usepackage{natbib}  % DO NOT CHANGE THIS AND DO NOT ADD ANY OPTIONS TO IT
\usepackage{caption} % DO NOT CHANGE THIS AND DO NOT ADD ANY OPTIONS TO IT
\usepackage{algorithm}
\usepackage{algorithmic}
\usepackage{graphicx}
\usepackage{subcaption}
\usepackage{graphicx}
\usepackage{amsmath}
\usepackage{amssymb}
\usepackage{algorithm}
\usepackage{algorithmic}
\usepackage{booktabs}
\usepackage{multirow}
\usepackage[table]{xcolor}
\usepackage{graphicx}
\usepackage{booktabs}
\usepackage[table]{xcolor}
\usepackage{threeparttable}
\usepackage{graphicx}

\definecolor{modelrow}{HTML}{EDF4F6} % low-saturation blue
\definecolor{bestrow}{HTML}{E8F2EA}  % low-saturation green
\usepackage{pifont}

\usepackage{newfloat}
\usepackage{listings}
\DeclareCaptionStyle{ruled}{labelfont=normalfont,labelsep=colon,strut=off} % DO NOT CHANGE THIS
\floatstyle{ruled}
\newfloat{listing}{tb}{lst}{}
\floatname{listing}{Listing}

\usepackage{booktabs}

\title{HB-VLA: Pushing 1-Bit Post-Training Quantization \\ for Vision-Language-Action Models}
\author{
    Xin Yan\textsuperscript{\rm 1},
    Zhenglin Wan\textsuperscript{\rm 2},
    Feiyang Ye,
    Xingrui Yu\textsuperscript{\rm 3}\thanks{Corresponding author.},
    Hangyu Du\textsuperscript{\rm 4},
    Yang You\textsuperscript{\rm 2},
    Ivor Tsang\textsuperscript{\rm 3}
}

\affiliations{
    \textsuperscript{\rm 1}School of Artificial Intelligence, Beijing Normal University, Beijing, China\\
    \textsuperscript{\rm 2}Department of Computer Science, National University of Singapore, Singapore\\
    \textsuperscript{\rm 3}Centre for Frontier AI Research, Agency for Science, Technology and Research (A*STAR), Singapore\\
    \textsuperscript{\rm 4}College of Design and Engineering, National University of Singapore, Singapore\\
    yu\_xingrui@a-star.edu.sg\\
}

\begin{document}

\maketitle

\begin{abstract}

Vision-Language-Action (VLA) models enable instruction-following embodied control, but their large compute and memory footprints hinder deployment on resource-constrained robots and edge platforms. While reducing weights to 1-bit precision through binarization can greatly improve efficiency, existing methods fail to preserve action-critical information under extreme compression, causing quantization errors to accumulate during long-horizon closed-loop execution and severely degrade control performance. To address this challenge, we propose HB-VLA, a VLA-tailored binarization framework. First, we introduce an action-aware rectified Hessian that identifies weights truly critical for stable action generation, where token importance is derived from a drift-weighted action loss. Then, we employ a sparse orthogonal transform for non-salient weights to induce a low-entropy intermediate state. Finally, we quantize both salient and non-salient weights in the Haar domain with group-wise 1-bit quantization. Extensive experiments on LIBERO, SIMPLER, and real-world robotic tasks show that HB-VLA reduces weight memory by about 82.0\%, while outperforming the strongest binary PTQ baseline by 11.0 percentage points on average across eight evaluation settings and by up to 17.8 points. These results demonstrate that HB-VLA enables efficient and stable deployment of VLA models on resource-limited robotic platforms.

\end{abstract}

\section{Introduction}
In recent years, Vision-Language-Action (VLA) models have emerged as a powerful paradigm for instruction-following embodied control by integrating visual perception, language understanding, and action generation within a single policy. Representative VLAs, such as RT-2~\citep{zitkovich2023rt}, OpenVLA~\citep{kim24}, and CogACT~\citep{li2024cogact}, have demonstrated strong performance on a wide range of manipulation tasks, including long-horizon instruction execution and generalization across diverse objects and scenes. These advances are largely enabled by large-scale model parameters and extensive robot datasets, which together provide high representation capacity and robust instruction grounding. However, these capabilities come with substantial computational demands and high memory usage, posing significant challenges for deployment on resource-constrained robotic platforms and edge devices where real-time, closed-loop control is required.
\begin{figure*}[!t]
  \centering
  \includegraphics[width=0.9\textwidth]{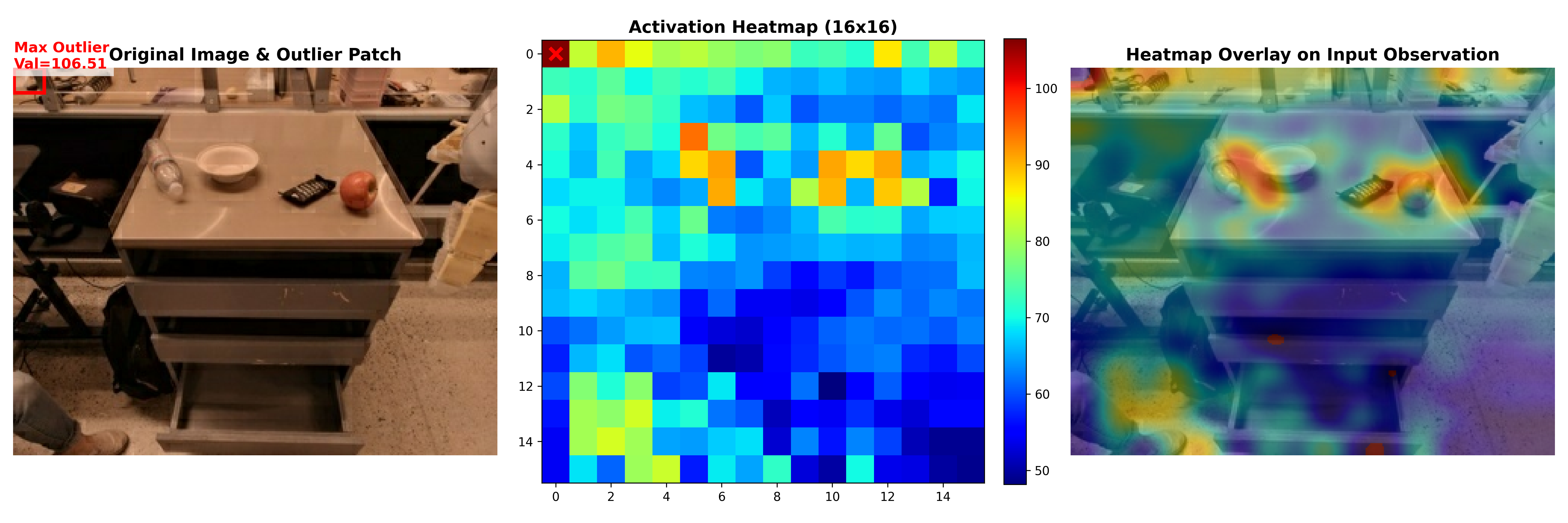}
  \caption{\textbf{Left:} An extreme background activation (\mbox{Val}=106.5).
\textbf{Middle:} The activation heatmap is dominated by statistical outliers.
\textbf{Right:} The overlay reveals misalignment: distractors such as the water bottle and background clutter dominate over the task-relevant apple, illustrating the dual-dominance problem.}
  \label{fig:motivation}
  \vskip -0.1in
\end{figure*}

Post-Training Quantization (PTQ)~\citep{nagel2019data,nagel2020up,krishnamoorthi2018quantizing} has gained significant traction due to its efficiency and practicality. Unlike quantization-aware training (QAT)~\citep{jacob2018quantization}, which requires access to training data and retraining the network, PTQ operates on frozen parameters and uses a small calibration set to determine quantization parameters and rounding decisions. Recent PTQ methods have shown promising results in reducing weight and activation bitwidths for large models \citep{lin2024awq}. Despite progress in 8-bit and 4-bit quantization for VLAs \citep{fang2025sqap}, the scale and deployment constraints of modern VLAs still call for more aggressive compression. Neural network binarization, which reduces weight precision to a single bit, is a promising direction toward ultra-low-bit quantization ($\leq$2 bits). However, directly transferring binary PTQ methods from large language models (LLMs) and vision-language models (VLMs) to VLAs is often ineffective. The central challenge in VLA binarization is not merely preserving feature reconstruction or language-model perplexity, but maintaining stable action behavior under closed-loop execution. Unlike LLMs or VLMs, whose errors are typically measured in token, logit, or feature space, VLA policies output continuous actions that are executed by a physical system. Consequently, even small quantization-induced action deviations can be amplified by contact dynamics and accumulated over long-horizon rollouts, leading to unstable grasps, oscillatory motions, or large trajectory drift. This fundamental mismatch suggests that ultra-low-bit VLA quantization should be formulated as an action-preserving problem, rather than a direct extension of reconstruction-oriented LLM/VLM binarization.

Motivated by this perspective, we revisit 1-bit PTQ for VLAs from the
viewpoint of policy preservation and identify a fundamental mismatch between
existing quantization criteria and embodied control requirements. Unlike
conventional models where quantization sensitivity can be approximated by
representation reconstruction, VLA policies require preserving the parameters
that are most influential to downstream action generation. This mismatch
manifests in three aspects. First, quantization sensitivity is inherently
policy-dependent and varies significantly across VLA components: while the
vision encoder is relatively robust, the projector and action generation
modules are substantially more sensitive due to their direct influence on
control outputs. Therefore, binary representation capacity should be allocated
according to action relevance rather than uniformly across the network.
Second, commonly used Hessian-based saliency estimation, e.g.,
$\mathbf{H}=\mathbf{X}\mathbf{X}^{\top}$, captures activation statistics but
does not account for the physical consequence of quantization errors. As shown
in Figure~\ref{fig:motivation}, high-magnitude visual activations caused by
background distractors can dominate the curvature estimation, leading to
saliency scores that are misaligned with action-critical regions. Third, the
parameter geometry of multimodal VLA policies introduces additional challenges
for extreme quantization. Existing transform-based binarization methods rely
on local structural assumptions in weight space, which may not hold when
parameters from different modalities are interleaved, resulting in inefficient
binary representations.

Based on these observations, we propose \textbf{HB-VLA}, an
action-preserving 1-bit post-training quantization framework that explicitly
optimizes binary representation allocation according to policy sensitivity.
Rather than extending reconstruction-oriented binarization methods to VLAs,
HB-VLA establishes a policy-aware quantization pipeline consisting of three
interconnected components. First, we introduce an action-aware rectified
Hessian to estimate the sensitivity of model parameters with respect to
closed-loop action deviation, enabling the identification of parameters whose
preservation is essential for stable control. Second, we develop a
structure-aware binary projection strategy that adapts the weight geometry of
multimodal policies before quantization, reducing the interference caused by
heterogeneous parameter distributions. Third, we employ a saliency-adaptive
Haar-domain quantization scheme that assigns different binary encoding
strategies to action-critical and redundant parameters while reducing storage
overhead. Through this unified design, HB-VLA transforms binary quantization
from a representation reconstruction problem into a policy preservation
problem, enabling efficient extreme compression while maintaining stable
embodied control.

% \begin{figure*}[!t]
% \centering
% \includegraphics[width=1.0\textwidth,height=0.4\textheight]{overview3.pdf}
% \vspace{-6pt}
% \caption{The pipeline of our \textbf{HBVLA} framework consists of two steps.}
% \label{fig:overview}
% \vspace{-10pt}
% \end{figure*}

\begin{figure*}[!t]
\centering
\includegraphics[width=0.86\textwidth]{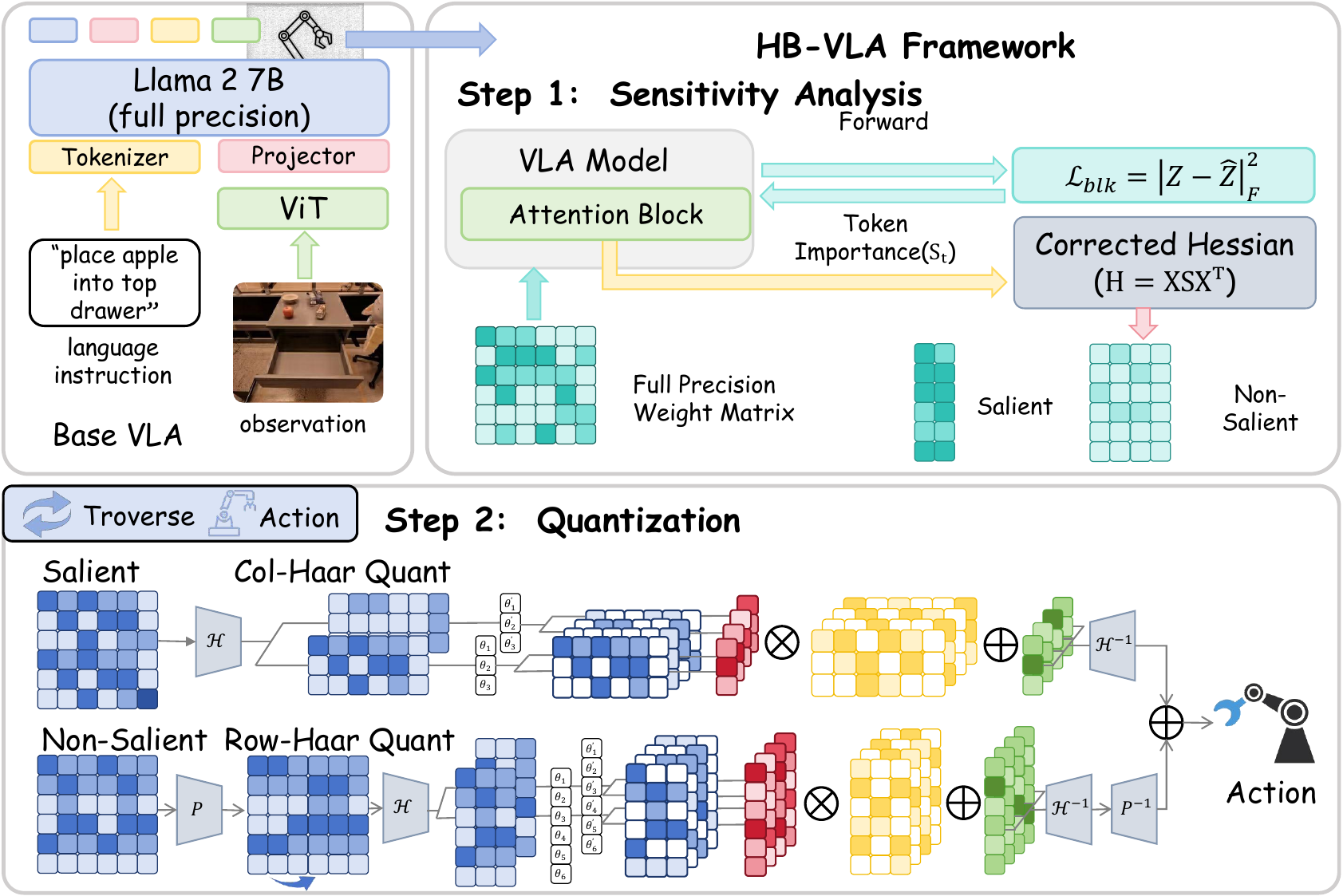}
\vspace{-6pt}
\caption{%The pipeline of our \textbf{HBVLA} framework consists of two steps: 
  (i) We establish a block-wise gradient probe to derive token importance scores ($S_t$) and construct a \textbf{Corrected Hessian} proxy to identify \textit{functionally salient} weights grounded in the policy. 
  (ii) We apply a hybrid quantization strategy: salient weights undergo residual quantization, while non-salient weights are processed via sparse orthogonal transform ($P$) and Haar wavelet transform ($\mathcal{H}$) prior to group-wise 1-bit quantization.
  %The pipeline of our \textbf{HBVLA} framework consists of two steps.
  }
\label{fig:overview}
\vspace{-10pt}
\end{figure*}

\section{Related Work}

\subsection{Vision-Language-Action Models.}
Vision-Language-Action (VLA) models have advanced embodied AI by enabling end-to-end policies that map multimodal observations to robotic actions. Representative models, including RT-2~\citep{zitkovich2023rt}, OpenVLA~\citep{kim24}, and UniVLA~\citep{bu2025univla}, extend pretrained VLMs~\citep{karamcheti2024prismatic} by tokenizing actions for sequence generation. In contrast, another line of work models continuous actions with generative decoders, such as diffusion policies in Octo~\citep{team2024octo} and CogACT~\citep{li2024cogact}, and flow matching in $\pi_0$~\citep{black2024pi0}, achieving smooth and precise control. However, their large memory footprint limits deployment. Existing quantization methods, such as BitVLA~\citep{wang2025bitvla} and SQIP~\citep{park2025saliency}, mainly rely on QAT, which requires costly training and large-scale robotic data. Therefore, low-bit PTQ for VLA models remains largely unexplored. This work addresses this gap by demonstrating that extreme quantization is essential for practical deployment of generalist VLA models.

\subsection{Network Binarization}
Network binarization compresses neural networks by constraining weights and activations to binary values, enabling efficient bitwise computation. Most existing approaches rely on QAT with Straight-Through Estimators (STE)~\citep{bengio2013estimating} to optimize the non-differentiable sign function. BinaryConnect~\citep{courbariaux2015binaryconnect} and BWN introduced weight binarization, while XNOR-Net~\citep{rastegari2016xnor} extended it to both weights and activations. For transformers~\citep{vaswani2017attention}, BiT~\citep{liu2022bit} introduced elastic binarization with learnable scaling, and EcoFormer~\citep{liu2022ecoformer} explored binary attention through hashing. Recent LLM binarization methods, including PB-LLM~\citep{yuan2024pbllm}, BiLLM~\citep{huang2024billm}, ARB-LLM~\citep{li2025arblm}, and HBLLM~\citep{chen2025hbllm}, improve binary representation capacity through various strategies, such as frequency decomposition. Bi-VLM~\citep{wang2025bi} further extends binarization to multimodal models but overlooks important activation structures. Inspired by these advances, we propose HB-VLA, an action-preserving 1-bit PTQ framework tailored for embodied control.

\section{Methodology}
\subsection{Method Overview}
 We define the objective of HB-VLA under the policy-preserving binary quantization setting. From a global perspective, we frame the quantization as an optimization problem to identify the optimal binary parameter configuration $\hat{\theta}^{\star}$ such that the induced policy $f_{\hat{\theta}}$ minimizes the KL divergence to the full-precision policy $f_{\theta}$:
\begin{equation}
\hat{\theta}^{\star} = \arg\min_{\hat{\theta}} \ \mathbb{E}_{x\sim\mathcal{D}} \left[ D_{\mathrm{KL}} \!\left( f_{\theta}(\cdot\mid x)\ \|\ f_{\hat{\theta}}(\cdot\mid x) \right) \right].
\label{eq:global_obj}
\end{equation}
For the quantization of a matrix layer $\mathbf{W}$, the objective expressed in the Frobenius norm is formulated as:
\begin{equation}
\min_{\widehat{\mathbf{W}}}\ \left\| \mathbf{W}\mathbf{X} - \widehat{\mathbf{W}}\mathbf{X} \right\|_F^2,
\end{equation}
where $\mathbf{X}$ is the input of the matrix layer. 

We emphasize that our approach does not aim to solve these objective functions via explicit end-to-end optimization. Instead, this formulation serves as a conceptual framework that guides our method design. The actual quantization process is based on a set of heuristics and structure-aware strategies that approximate these objectives in a computationally efficient manner. To this end, our method HB-VLA adapts the efficient Haar transform into a unified pipeline tailored for VLAs. As shown in Figure~\ref{fig:overview}, we begin by computing a policy-aware Hessian to identify salient weights critical for action. For the remaining non-salient weights, we apply a structured sparse orthogonal transform prior to the Haar transform. This aligns weight columns into a low-entropy state to suppress high-frequency noise. Subsequently, we perform \textbf{residual-aware refinement}, where salient columns are quantized via a column Haar transform to compensate for non-salient approximation errors.

\subsection{Policy-Aware Weight Partitioning}

For each quantized linear layer, let $\mathbf{X}=[\mathbf{x}_1,\ldots,\mathbf{x}_N]\in\mathbb{R}^{d\times N}$ denote the calibration activations, where $N$ is the number of input tokens. 
Given a weight matrix $\mathbf{W}$, our goal is to identify a compact set of salient columns whose binarization errors have the largest impact on action generation. 
Instead of using the standard Hessian proxy $\mathbf{H}=\mathbf{X}\mathbf{X}^{\top}$, we introduce a rectified Hessian that assigns different importance to different tokens:
\begin{equation}
\tilde{\mathbf{H}} \triangleq \mathbf{X}\mathbf{S}\mathbf{X}^{\top}
= \sum_{t=1}^{N}s_t\mathbf{x}_t\mathbf{x}_t^{\top},
\label{eq:rectified_hessian}
\end{equation}
where $\mathbf{S}=\mathrm{Diag}(s_1,\ldots,s_N)$ and $s_t$ is the unique token-importance score defined in Eq.~\eqref{eq:token_importance} from the drift-weighted action loss. 
This formulation preserves the efficiency of Hessian-based saliency estimation while aligning the weighting term with action-space sensitivity.

\textbf{Forward.}
We consider a local residual attention block in the action pathway,
$\Phi(\mathbf{X}) \triangleq \mathbf{X}+\mathrm{MHSA}(\mathbf{X})$,
and its binarized counterpart $\hat{\Phi}(\cdot)$.
Given the same input $\mathbf{X}\in\mathbb{R}^{d\times N}$, we obtain
\begin{equation}
\mathbf{Z}=\Phi(\mathbf{X}),\qquad 
\hat{\mathbf{Z}}=\hat{\Phi}(\mathbf{X}).
\end{equation}
The two block outputs are then passed through the remaining frozen layers to produce the full-precision action $a$ and the binarized action $\hat{a}$, respectively.

\textbf{Drift-Weighted Action Loss.}
Instead of measuring only local block reconstruction error, we measure the action deviation induced by binarization.
For a $d_a$-dimensional action vector, we define a drift-weighted action loss:
\begin{equation}
\mathcal{L}_{\mathrm{act}} =
\sum_{j=1}^{d_a}
\bar{\rho}_j
\left(\hat{a}_j-a_j\right)^2,
\end{equation}
where $\bar{\rho}_j$ denotes the drift sensitivity of the $j$-th action dimension.
Following the idea of structural kinematic propagation, we estimate this sensitivity using a structural or embodiment-specific Jacobian that maps action perturbations to end-effector deviations:
\begin{equation}
\rho_j =
\mathbb{E}_t
\left[
\left\|
\mathbf{J}^{(t)}_{:,j}
\right\|_2
\right],
\qquad
\bar{\rho}_j =
\frac{\rho_j}
{\frac{1}{d_a}\sum_{k=1}^{d_a}\rho_k+\epsilon}.
\end{equation}
Here, $\mathbf{J}^{(t)}$ describes how an error in each action dimension affects the end-effector at timestep $t$.
Thus, action dimensions whose perturbations are more likely to be amplified into trajectory drift receive larger weights in $\mathcal{L}_{\mathrm{act}}$.

\textbf{Backward.}
We perform a locally isolated backpropagation on $\mathcal{L}_{\mathrm{act}}$ and collect gradients at the attention projections $(Q,K,V,O)$.
Let $\mathcal{P}=\{Q,K,V,O\}$ denote the set of projection outputs
$\mathbf{Y}^{(p)}\in\mathbb{R}^{d_p\times N}$.
The projection gradients and the resulting token-importance score are defined jointly as
\begin{equation}
\begin{aligned}
\mathbf{G}^{(p)}
&\triangleq \frac{\partial \mathcal{L}_{\mathrm{act}}}
{\partial \mathbf{Y}^{(p)}}
\in \mathbb{R}^{d_p\times N},\quad p\in\mathcal{P},\\
a_t^{(p)}
&\triangleq \frac{1}{d_p}\bigl\|\mathbf{G}^{(p)}_{:,t}\bigr\|_2,\\
s_t
&\triangleq \frac{1}{|\mathcal{P}|}
\sum_{p\in\mathcal{P}} a_t^{(p)},\quad t=1,\ldots,N.
\end{aligned}
\label{eq:token_importance}
\end{equation}
Here, $a_t^{(p)}$ is the channel-normalized sensitivity of token $t$ at projection $p$, and $s_t$ averages this sensitivity over $Q$, $K$, $V$, and $O$. A larger $s_t$ therefore indicates that perturbing token $t$ causes a larger drift-weighted action deviation across the attention pathways. Equation~\eqref{eq:token_importance} is the sole definition of token importance used in both the main text and the appendix; no separate residual-feature gradient is used as an alternative score.

\textbf{Saliency Partition.}
Finally, we compute column saliency from the rectified Hessian $\tilde{\mathbf{H}}$ in Eq.~\eqref{eq:rectified_hessian} and partition each layer into salient and non-salient column sets $\mathcal{I}_{\mathrm{sal}}$ and $\mathcal{I}_{\mathrm{non-sal}}$. 
We first derive element-wise importance scores normalized by the Hessian diagonal and aggregate them into per-column scores via an $\ell_2$ reduction. 
The number of salient columns is then determined by minimizing a local binarization reconstruction objective, with the remaining columns assigned to $\mathcal{I}_{\mathrm{non-sal}}$.

\subsection{Saliency-Aware Hybrid Binarization With Haar Transform}
We use a hybrid binarization method that treats salient and non-salient weights differently, but both parts share the same two-step quantization primitive. We first map weights to the Haar domain and split them into low-pass and high-pass subbands.
Concretely, for an even length $m$, we define a one-level, orthonormally normalized Haar transform matrix
$\mathbf{H}_m\in\mathbb{R}^{m\times m}$ and map $w\in\mathbb{R}^{1\times m}$ to the Haar domain by
\begin{equation}
\begin{aligned}
\mathcal{H}(w)\ \triangleq\ w\mathbf{H}_m
&=\big[w^{\mathrm{lo}},\,w^{\mathrm{hi}}\big],\\[-2pt]
w_k^{\mathrm{lo}}&=\frac{w_{2k-1}+w_{2k}}{\sqrt{2}},\\[-2pt]
w_k^{\mathrm{hi}}&=\frac{w_{2k-1}-w_{2k}}{\sqrt{2}}.
\end{aligned}
\label{eq:haar_def}
\end{equation}
Here $k=1,\ldots,m/2$, and $w^{\mathrm{lo}},w^{\mathrm{hi}}\in\mathbb{R}^{1\times (m/2)}$ are the low-pass and high-pass subbands. Thus, $\mathbf{H}_m$ is constructed from the normalized filters $[1/\sqrt{2},1/\sqrt{2}]$ and $[1/\sqrt{2},-1/\sqrt{2}]$ (with a coefficient-ordering permutation), so that $\mathbf{H}_m^{\top}\mathbf{H}_m=\mathbf{H}_m\mathbf{H}_m^{\top}=\mathbf{I}$ and $\mathbf{H}_m^{-1}=\mathbf{H}_m^{\top}$.
More details are provided in Appendix~\ref{app:haar_highpass_identity}.

After the transform, we employ a row-wise grouping strategy. Instead of global partitioning, we adaptively split coefficients within each frequency band into groups (e.g., dense/sparse) to capture the local structural pattern. Then we apply binarization to every group:
\begin{equation}
Q(u)=\alpha_g\cdot \mathrm{sign}(u-\mu_g),
\label{eq:centered_sign}
\end{equation}
where $\mu_g$ and $\alpha_g$ are computed per group. Notably, for non-salient weights, we optimize storage efficiency by enforcing a single shared mean $\mu$ across groups within the same row and frequency band. This approach effectively reduces the average bit-width and metadata overhead while preserving model performance.

\textbf{Non-salient Weights Binarization.}
Since salient columns are excluded from the Haar transform of the non-salient part, we first fill the
missing values in salient columns using adjacent averages, obtaining $\mathbf{W}_{l,\mathrm{filled}}$.
We binarize $\mathbf{W}_{l,\mathrm{filled}}$ in the Haar domain by applying a row-wise Haar transform along the
column axis:
\begin{equation}
\mathbf{W}^{c}_{l,\mathrm{non\text{-}sal}}
\;=\;
\mathcal{H}_{\text{row}}\!\left(\mathbf{W}_{l,\mathrm{filled}}\right)
\;=\;
\mathbf{W}_{l,\mathrm{filled}}\,\mathbf{H}_m.
\label{eq:row_haar_nonsal}
\end{equation}
In VLAs, non-salient columns often exhibit strong modality-dependent structures. However, these columns are typically interleaved in the weight matrix space rather than forming contiguous, modality-specific blocks. This lack of structural locality poses a significant challenge for the Haar transform, which operates on fixed local windows by calculating differences between adjacent columns. When the Haar operator pairs columns from disparate modalities, the resulting cross-modality offsets manifest as sharp step-change outliers in the Haar domain.

To resolve this, we employ a \textbf{sparse orthogonal transform}, implemented via a permutation matrix $\mathbf{P} \in \{0, 1\}^{m \times m}$, prior to the Haar transform. This mechanism allows us to maintain the $\mathcal{O}(d)$ convolutional efficiency of Haar while making it adaptive to the underlying weight geometry. After permutation, we follow the quantization primitive:
\begin{align}
\mathbf{U} &= \mathbf{W}_{l,\mathrm{non-sal}}\mathbf{P}\mathbf{H}_m, \notag \\
\mathbf{U}_{B} &= Q(\mathbf{U}) = \boldsymbol{\alpha}\odot \mathrm{sign}\!\big(\mathbf{U}-\boldsymbol{\mu}\big), \label{eq:unsalient_pipeline} \\
\widehat{\mathbf{W}}_{l,\mathrm{non-sal}} &= \mathbf{U}_{B}\mathbf{H}_m^{\top}\mathbf{P}^{\top}, \notag
\end{align}
where $\boldsymbol{\mu}$ and $\boldsymbol{\alpha}$ are computed group-wise in Eq.~\eqref{eq:centered_sign}. Since both the permutation $\mathbf{P}$ and the normalized Haar basis $\mathbf{H}_m$ are orthogonal, the transform preserves the Frobenius norm. In particular,
\begin{equation*}
\begin{aligned}
&\|\mathbf{W}_{l,\mathrm{non-sal}}-\widehat{\mathbf{W}}_{l,\mathrm{non-sal}}\|_{F}\\[-2pt]
&\qquad=\|\mathbf{U}-\mathbf{U}_B\|_{F}.
\end{aligned}
\end{equation*}

We construct this sparse orthogonal transform by minimizing the high-frequency energy generated during decomposition. Specifically, for a one-level Haar transform, the energy in the high-pass subband is proportional to the sum of discrepancies between paired columns. 
Let $\pi$ be the ordering induced by $\mathbf{P}$ and let $\mathbf{H}_{\mathrm{hi}}$ denote the first-level high-pass basis.
Then the high-pass energy admits the following identity:
\begin{equation}
\begin{split}
\big\|\mathbf{W}_{l,\mathrm{non\text{-}sal}}\mathbf{P}\mathbf{H}_{\mathrm{hi}}\big\|_F^2 = & \frac{1}{2}\sum_{k=1}^{\lfloor m/2\rfloor} \big\| \mathbf{W}_{l,\mathrm{non\text{-}sal}}(:,\pi(2k-1)) \\
& - \mathbf{W}_{l,\mathrm{non\text{-}sal}}(:,\pi(2k)) \big\|_2^2,
\end{split}
\label{eq:hf_energy_identity}
\end{equation}
and we provide a proof in Appendix~\ref{app:haar_highpass_identity}.
Therefore, determining the optimal $\mathbf{P}$ is equivalent to solving a discrete optimization problem that maximizes local column similarity. Since the exact global optimum is intractable, we adopt a greedy pairing-and-chaining heuristic to efficiently generate the permutation $\pi$ that defines $\mathbf{P}$.
\textit{Due to the page limitation, the detailed pseudocode of the proposed algorithm is provided in the Appendix. }

\textbf{Salient Weights Binarization.}
After obtaining the non-salient approximation $\widehat{\mathbf{W}}_{l,\mathrm{non-sal}}$, we quantize salient columns
on the residual to avoid interference from the non-salient reconstruction. We define the residual
\begin{equation}
\mathbf{R}_l \;=\; \mathbf{W}_l - \widehat{\mathbf{W}}_{l,\mathrm{non-sal}}.
\label{eq:residual_def}
\end{equation}
By defining $\mathcal{I}_{\mathrm{sal}}$ as the index set of salient columns, we extract the salient residual
$\mathbf{R}_l(:,\mathcal{I}_{\mathrm{sal}})$ and apply a \emph{column-wise} Haar transform (more details are provided in Appendix~\ref{app:haar_colwise}):
\begin{equation}
\mathbf{R}^{c}_{l,\mathrm{sal}}
\;=\;
\mathbf{H}_n\,\mathbf{R}_l(:,\mathcal{I}_{\mathrm{sal}}),
\label{eq:col_haar_sal}
\end{equation}
where $\mathbf{R}^{c}_{l,\mathrm{sal}}$ denotes the Haar-domain coefficients of the salient residual. 
We then apply the same quantization primitive in the Haar domain using a column-wise transform, followed by the inverse transform:
\begin{equation}
\widehat{\mathbf{W}}_{l,\mathrm{sal}}
\;=\;
\mathcal{H}_{\mathrm{col}}^{-1}\!\Big( Q_(\mathbf{R}^{c}_{l,\mathrm{sal}}) \Big)
\;=\;
\mathbf{H}_n^{-1}\, Q_(\mathbf{R}^{c}_{l,\mathrm{sal}}).
\label{eq:sal_quant_refined}
\end{equation}
Finally, we reconstruct the quantized layer weights by summing the two components: 
\begin{equation}
\widehat{\mathbf{W}}_l
\;=\;
\widehat{\mathbf{W}}_{l,\mathrm{non-sal}} \;+\; \widehat{\mathbf{W}}_{l,\mathrm{sal}}.
\label{eq:final_recon}
\end{equation}

\section{Experiment}
\label{sect:exp}
 We validate our approach through a series of experiments across simulated benchmarks and physical robot manipulation tasks. All experiments are conducted on NVIDIA A800 GPUs.  More details are provided in  Appendix.
 
\subsection{Experimental Setup}
\label{sect:setup}
We conduct experiments on three settings: the LIBERO \citep{liu2023} and SIMPLER \citep{li2024}, two widely recognized robotic manipulation benchmarks for evaluation in simulation environments, and the real-world Mobile ALOHA platform with three test tasks, to validate the practical applicability of HB-VLA. We use the widely used performance evaluation metrics ``Success Rate (SR)” to evaluate the results in these challenging settings. During inference, all activations remain in BF16 precision, while only weights are binarized. 

\noindent \textbf{LIBERO}~\citep{liu2023}: LIBERO evaluates knowledge transfer and policy generalization across four suites: Spatial, Object, Goal, and Long. It contains 100 MuJoCo tasks and 5,000 demonstrations using a Franka Panda arm with RGB, proprioceptive, and delta-action observations.

\noindent \textbf{SIMPLER}~\citep{li2024}: SIMPLER evaluates VLA policies in high-fidelity simulations under Visual Matching and Variant Aggregation. We use the Google robot arm on four tasks: Pick Coke Can, Move Near, Open/Close Drawer, and Place Apple into Top Drawer.

\noindent \textbf{Real-world Evaluation Suite.} We evaluate real-world adaptability on Mobile ALOHA using Pick and Place, Sequenced Instruction, and Flexible Folding benchmarks~\citep{bu2025univla, zhao2025cot}. These tasks assess generalization from 30--450 demonstrations per task to novel environments. Further details are provided in the Appendix.
%\ref{app:real}

\noindent \textbf{Baselines.} We selected recently published 1-bit PTQ methods as baselines, including HBLLM \citep{chen2025hbllm}, BiLLM \citep{huang2024billm}, and BiVLM \citep{wang2025bi}. BiLLM and HBLLM all utilize the PTQ approach for model calibration through OBQ based method of GPTQ.

\begin{table*}[t]
\centering
\small
\renewcommand{\arraystretch}{1.20}
\setlength{\tabcolsep}{4.5pt}
\resizebox{0.85\textwidth}{!}{%
\begin{tabular}{c l c c c c c c c c}
\toprule
SIMPLER
& Task
& FP
& Vis
& Adp
& Lm
& Act
& Vis+Lm
& Vis+Adp+Lm
& Vis+Adp+Lm+Act \\
\midrule

\multirow{5}{*}{\shortstack{Visual\\Matching}}
& Pick Coke
& 91.3 & 90.8 & 84.7 & 88.2 & 85.4 & 86.7 & 87.2 & 86.4 \\

& Move Near
& 85.0 & 84.6 & 82.7 & 83.5 & 82.5 & 81.7 & 80.3 & 79.7 \\

& O/C Drawer
& 71.8 & 72.0 & 70.6 & 71.5 & 70.0 & 71.8 & 69.9 & 68.3 \\

& Place Apple
& 50.9 & 47.4 & 46.9 & 44.8 & 41.7 & 38.4 & 36.2 & 34.5 \\

& \cellcolor{gray!10}Avg
& \cellcolor{gray!10}74.8
& \cellcolor{gray!10}73.7
& \cellcolor{gray!10}71.2
& \cellcolor{gray!10}72.0
& \cellcolor{gray!10}69.9
& \cellcolor{gray!10}70.0
& \cellcolor{gray!10}68.4
& \cellcolor{gray!10}67.2 \\
\midrule
\multirow{5}{*}{\shortstack{Variant\\Aggregation}}
& Pick Coke
& 89.6 & 89.3 & 86.9 & 87.4 & 82.7 & 85.1 & 84.9 & 82.5 \\
& Move Near
& 80.8 & 80.8 & 75.8 & 76.3 & 74.4 & 72.5 & 73.2 & 71.7 \\
& O/C Drawer
& 28.3 & 26.7 & 25.9 & 25.7 & 24.3 & 21.4 & 20.6 & 20.3 \\
& Place Apple
& 46.6 & 45.4 & 30.4 & 32.7 & 29.5 & 24.9 & 25.3 & 22.6 \\
& \cellcolor{gray!10}Avg
& \cellcolor{gray!10}61.3
& \cellcolor{gray!10}60.6
& \cellcolor{gray!10}54.8
& \cellcolor{gray!10}55.5
& \cellcolor{gray!10}52.7
& \cellcolor{gray!10}51.0
& \cellcolor{gray!10}51.0
& \cellcolor{gray!10}49.2 \\
\bottomrule
\end{tabular}%
}
\vskip -0.05in
\caption{
Component-wise quantization results of CogACT model on SIMPLER environment, with average
weight precision ranging from 1.02 to 1.13 bits.
FP denotes full precision; Vis denotes the vision encoder;
Adp denotes the adaptation layer; Lm denotes the language model;
and Act denotes the action head.  “O/C Drawer” refers to the Open/Close Drawer task.
All results are reported as success rates (\%).
}
\label{tab:cogact_sim_components}
\vskip -0.1in
\end{table*}

\begin{table*}[t]
\centering
\renewcommand{\arraystretch}{1.25}
\setlength{\tabcolsep}{8pt}

\resizebox{0.8\textwidth}{!}{%
\begin{tabular}{lcccccccc}
\toprule
LIBERO & FP & Vis & Adp & Lm & Act
& Vis+Lm & Vis+Adp+Lm & Vis+Adp+Lm+Act \\
\midrule
Spatial & 98.5 & 98.5 & 97.6 & 97.9 & 96.5 & 95.8 & 94.2 & 92.8 \\
Object  & 99.0 & 98.6 & 98.0 & 98.2 & 97.8 & 97.2 & 96.5 & 95.6 \\
Goal    & 97.5 & 97.3 & 96.2 & 95.7 & 91.3 & 93.5 & 89.8 & 86.5 \\
Long    & 93.5 & 92.7 & 87.3 & 89.5 & 86.5 & 84.4 & 79.2 & 76.7 \\
\cellcolor{gray!10} Avg     & \cellcolor{gray!10}97.1 &\cellcolor{gray!10} 96.8 &\cellcolor{gray!10} 94.8 &\cellcolor{gray!10} 95.3 &\cellcolor{gray!10} 93.0 &\cellcolor{gray!10} 92.7 &\cellcolor{gray!10} 89.9 &\cellcolor{gray!10} 87.9 \\
\bottomrule
\end{tabular}%
}
\vskip -0.05in
\caption{Quantization on different components of $\pi_{0.5}$ on LIBERO
with weights from 1.02 to 1.13 bits. All results are reported as success rates (\%).
FP: Full precision. Vis: Vision encoder. Adp: Adapt layer.
Lm: Language model. Act: Action Head.}
\label{tab:pi05_libero_components}
\vskip -0.1in
\end{table*}

\begin{table*}[t]
\centering
\vspace{0.3em}

\renewcommand{\arraystretch}{1.25}
\setlength{\tabcolsep}{8pt}

\resizebox{0.8\textwidth}{!}{%
\begin{tabular}{lcccccccc}
\toprule
LIBERO & FP & Vis & Adp & Lm & 
Vis+Adp & Vis+Lm & Lm+Adp & Vis+Adp+Lm \\
\midrule
Spatial & 97.6 & 97.2 & 94.5 & 94.7 
& 90.8 & 89.3 & 90.2 & 83.4 \\

Object  & 98.4 & 97.6 & 96.8 & 97.2 
& 97.2 & 97.8 & 94.2 & 88.6 \\

Goal    & 94.5 & 93.9 & 93.4 & 92.8 
& 91.9 & 91.3 & 92.0 & 85.8 \\

Long    & 97.9 & 95.4 & 86.9 & 87.5 
& 85.5 & 82.7 & 86.2 & 76.0 \\

\cellcolor{gray!10}Avg     &\cellcolor{gray!10} 97.1 &\cellcolor{gray!10} 96.0 &\cellcolor{gray!10} 92.9 &\cellcolor{gray!10} 93.0 
&\cellcolor{gray!10} 91.4 &\cellcolor{gray!10} 90.3 &\cellcolor{gray!10} 90.7 &\cellcolor{gray!10} 83.5 \\
\bottomrule
\end{tabular}%
}
\vskip -0.1in
\caption{Quantization on different components of OpenVLA-OFT on LIBERO
with weights from 1.02 to 1.13 bits. All results are reported as success rates (\%).
FP: Full precision. Vis: Vision encoder. Adp: Adapt layer.
Lm: Language model.}
\label{tab:openvla_oft_libero_components}
\vskip -0.1in
\end{table*}

\begin{table*}[t]
\centering
\renewcommand{\arraystretch}{1.25}
\setlength{\tabcolsep}{8pt}

\resizebox{0.85\textwidth}{!}{\begin{tabular}{llccccc}
\toprule
Model & Benchmark & FP & BiLLM & BiVLM & HBLLM & \textbf{HB-VLA (Ours)} \\
\midrule
$\pi_{0.5}$ (LM+ViT) & LIBERO
& 97.1 & 54.3 & 63.2 & 74.9 & \textbf{92.7} \\

$\pi_{0.5}$ (All) & LIBERO
& 97.1 & 47.2 & 55.8 & 70.7 & \textbf{87.9} \\

CogACT (LM+ViT) & SIMPLER (VM)
& 74.8 & 28.8 & 57.1 & 62.3 & \textbf{70.0} \\

CogACT (All) & SIMPLER (VM)
& 74.8 & 23.2 & 50.7 & 55.8 & \textbf{67.2} \\

CogACT (LM+ViT) & SIMPLER (VA)
& 61.3 & 14.9 & 47.4 & 47.9 & \textbf{51.0} \\

CogACT (All) & SIMPLER (VA)
& 61.3 & 8.8 & 40.9 & 43.2 & \textbf{49.2} \\

OpenVLA-OFT (LM+ViT) & LIBERO
& 97.1 & 57.7 & 64.0 & 79.2 & \textbf{90.3} \\

OpenVLA-OFT (All) & LIBERO
& 97.1 & 49.3 & 52.5 & 70.1 & \textbf{83.5} \\
\bottomrule
\end{tabular}}

\vskip -0.1in
\caption{
SOTA comparison on $\pi_{0.5}$, CogACT, and OpenVLA-OFT
with weights ranging from 1.02 to 1.13 bits.
FP: Full precision. LM: Language model. ViT: Vision backbone.
All: The whole VLA model.
All results are reported as success rates (\%).
\textbf{Bold} values indicate the best results among the quantized models,
excluding the full-precision model.
}
\label{tab:compare}
\vskip -0.1in
\end{table*}

\subsection{Main Results}

\noindent \textbf{Quantization on Different Components.}
As shown in Tables~\ref{tab:cogact_sim_components},
\ref{tab:pi05_libero_components}, and
\ref{tab:openvla_oft_libero_components}, the vision encoder is generally
the most robust to quantization. The adaptor/projector and language
model show moderate sensitivity, while quantizing the action head causes
the largest performance drop.

\noindent \textbf{Comparison with SOTA Methods.}
As shown in Table~\ref{tab:compare}, HB-VLA consistently outperforms all
binary PTQ baselines on both LIBERO and SIMPLER. Existing LLM/VLM
binarization methods degrade substantially when directly applied to VLAs,
whereas HB-VLA retains a much smaller gap to full precision. On LIBERO,
HB-VLA reaches 92.7\% with LM+ViT quantization, only 4.4 percentage
points below FP. On SIMPLER Visual Matching, it improves the strongest
baseline from 62.3\% to 70.0\% for LM+ViT quantization and from 55.8\%
to 67.2\% for full-model quantization. Similar gains under Variant
Aggregation indicate better preservation of action-relevant
representations under visual shifts and closed-loop execution.

\begin{figure}[!t]
\centering
\includegraphics[width=\linewidth]{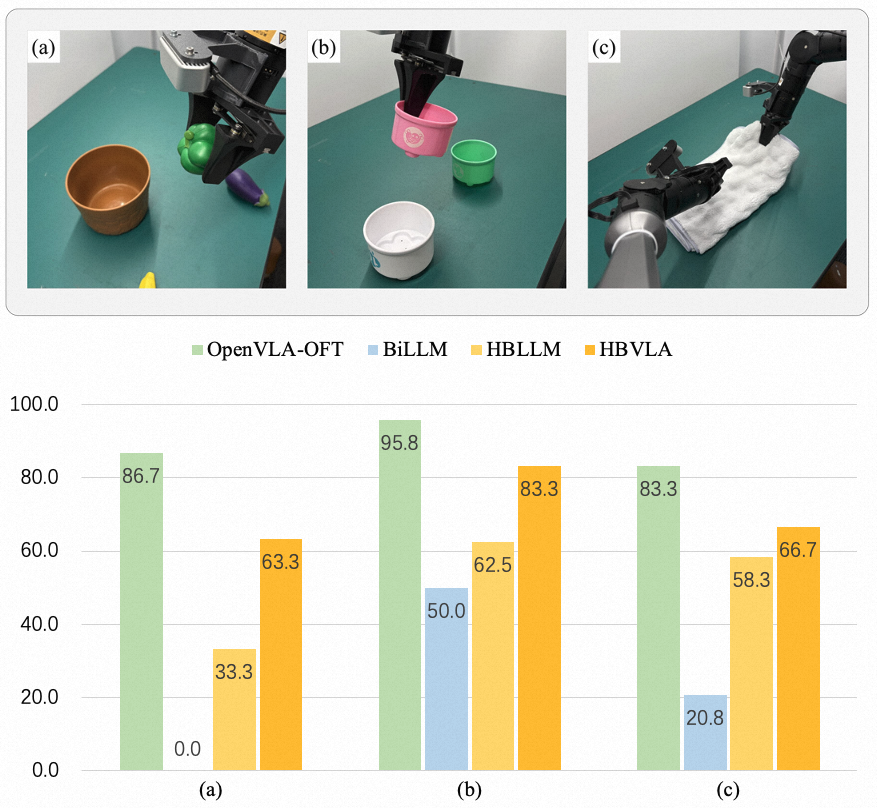}
\vskip -0.1in
\caption{Comparison of baseline methods on the Mobile ALOHA experiments. Evaluation across three real-world tasks, including (a) Pick and Place, (b) Sequenced Instruction, (c) Flexible Folding. \textbf{Top:} Middle state image for each task. \textbf{Bottom:} Task-specific success rates for OpenVLA-OFT (FP Model), our HBVLA method, and baselines, including BiLLM and HBLLM.
}
\label{fig:mobile_aloha_results}
\vskip -0.2in
\end{figure}

\noindent \textbf{Mobile ALOHA.}
Pick and Place is evaluated over 30 trials, while the other tasks use
24 trials each. As shown in Figure~\ref{fig:mobile_aloha_results},
OpenVLA-OFT performs strongly across all three tasks, whereas BiLLM
performs worst and exhibits persistent arm oscillations during execution.
Compared with FP, HB-VLA drops by 23.4, 12.5, and 16.6 percentage points
on Pick and Place, Sequenced Instruction, and Flexible Folding,
respectively. Despite these gaps, it consistently outperforms all binary
PTQ baselines, demonstrating better retention of real-world control
performance under 1-bit weight quantization.

\subsection{Time and Memory Analysis}
As shown in Table~\ref{tab:memory_comparison}, HB-VLA substantially reduces the weight memory footprint across different VLA models. 
Using the metadata-aware storage estimate, HB-VLA compresses $\pi_{0.5}$ from 4.6GB to 0.83GB, OpenVLA-OFT from 15.2GB to 2.74GB, and CogACT from 30.5GB to 5.50GB, corresponding to an average memory reduction of about 82.0\%. 
This demonstrates that HB-VLA remains storage-efficient even when quantization metadata is considered. 
In addition, HB-VLA achieves a $2.93\times$ inference speedup over the full-precision baseline, reducing latency by approximately 65.9\%. 
These results show that HB-VLA provides both substantial memory savings and practical acceleration for closed-loop robotic control.

\begin{table}[t]
\centering
% \resizebox{0.48\columnwidth}{!}{
\begin{tabular}{lcc}
\toprule
\textbf{Model} & \textbf{FP Memory} & \textbf{Quantization Memory} \\
\midrule
$\pi_{0.5}$ & 4.6GB  & 0.83GB \\
OpenVLA-OFT & 15.2GB & 2.74GB \\
CogACT & 30.5GB & 5.50GB \\
\bottomrule
\end{tabular}
% }
\caption{Weight-only memory comparison before and after HB-VLA quantization.}
\label{tab:memory_comparison}
\vskip -0.1in
\end{table}

\subsection{Further Analysis}
\noindent \textbf{Column-Norm Criterion for Permutation.} 
To evaluate the impact of the permutation criterion for non-salient columns on quantization effectiveness, we compare two strategies: the column $\ell_1$-norm and the column $\ell_2$-norm with CogACT on Simpler. The results are shown in Table~\ref{tab:l1}. Using the $\ell_2$-norm yields lower quantization error and superior downstream performance compared to $\ell_1$, as the $\ell_2$-norm more effectively captures energy distribution across columns, thereby improving quantization quality.

\begin{table}[t]
\centering
\renewcommand{\arraystretch}{1.15}
\setlength{\tabcolsep}{8pt}

\begin{tabular}{ccc}
\toprule
\shortstack{Selection\\criterion}
&
\shortstack{Visual\\Matching$\downarrow$}
&
\shortstack{Variant\\Aggregation$\downarrow$} \\
\midrule
$\ell_1$
& 11.6\%
& 15.6\% \\

$\ell_2$
& \textbf{8.8\%}
& \textbf{12.8\%} \\
\bottomrule
\end{tabular}
\caption{Study of column permutation criterion.}
\label{tab:l1}
\vskip -0.1in
\end{table}

\noindent \textbf{Effectiveness of Action-Aware Hessian.}
To investigate the impact of the proposed rectified Hessian on saliency estimation, we compare it against the standard Hessian formulation with CogACT on Simpler. 
As shown in Table~\ref{tab:hessian}, our action-aware Hessian effectively filters out quantization noise and accurately identifies task-critical weights essential for instruction-following control, resulting in significant gains in success rate. 

\begin{table}[t]
\centering
\renewcommand{\arraystretch}{1.15}
\setlength{\tabcolsep}{7pt}

\begin{tabular}{ccc}
\toprule
\shortstack{Hessian\\Formulation}
&
\shortstack{Visual\\Matching (Avg.)$\downarrow$}
&
\shortstack{Variant\\Aggregation (Avg.)$\downarrow$} \\
\midrule

Standard
& 12.5\%
& 13.4\% \\

Action-Aware
& \textbf{10.3\%}
& \textbf{12.1\%} \\

\bottomrule
\end{tabular}
\caption{Impact of Hessian formulation on quantization.}
\label{tab:hessian}
\vskip -0.1in
\end{table}

\textit{Due to the page limitation, additional ablation studies on saliency partition candidates and sparse orthogonal reordering are provided in the Appendix.}

\section{Conclusion}
In this paper, we introduce HB-VLA, a 1-bit post-training
quantization framework tailored to VLA models. HB-VLA combines
policy-aware weight partitioning with a sparse orthogonal transform
in the Haar domain to alleviate the severe performance degradation
caused by extreme quantization while preserving action-relevant
representations. Experiments on LIBERO, SIMPLER, and real-world
Mobile ALOHA demonstrate that HB-VLA consistently outperforms
existing binary PTQ baselines across different VLA architectures,
visual variations, and manipulation tasks. These results show that
HB-VLA achieves substantial model compression while retaining strong
closed-loop control performance, highlighting the practical potential
of low-bit PTQ for deploying capable VLA policies on
resource-constrained robotic platforms.

\clearpage

\bibliography{aaai2027}

\clearpage


\end{document}